\documentclass[11pt]{article}

\usepackage[review]{acl}
\usepackage{times}
\usepackage{latexsym}
\usepackage{float} 

\usepackage{titlesec}
\titlespacing*{\paragraph}{1pt}{1pt}{1em}
\nolinenumbers
\usepackage[T1]{fontenc}
\usepackage[utf8]{inputenc}
\usepackage{microtype}
\usepackage{inconsolata}

\usepackage{amsthm}

\usepackage{amsmath}
\usepackage{graphicx}
\usepackage{algorithm}
\usepackage[noend]{algpseudocode}
\usepackage{afterpage}

\usepackage{enumitem}

\usepackage{multirow}
\usepackage{adjustbox}
\usepackage{bbm}
\usepackage{wrapfig,lipsum}
\usepackage{xspace}
\usepackage{booktabs,cellspace}  
\usepackage[normalem]{ulem}
\usepackage{makecell}
\usepackage{amssymb}
\usepackage{pifont}
\usepackage{todonotes}
\newcommand{\xmark}{\ding{55}}%

\usepackage{microtype}

\usepackage{cleveref}

\usepackage{etoolbox}
\usepackage[tikz]{bclogo}

\usepackage{subcaption}
\usepackage{adjustbox}

\definecolor{msftBlue}{RGB}{0,164,239}
\definecolor{msftGreen}{RGB}{127,186,0}
\definecolor{msftYello}{RGB}{255,185,0}
\definecolor{msftBlack}{RGB}{0,0,0}

\usepackage{tcolorbox} 
\tcbuselibrary{skins} 
\usepackage[T1]{fontenc}

\tcbset{
    userstyle/.style={
        enhanced,
        colback=white,
        colframe=black,
        colbacktitle=gray!20,
        coltitle=black,
        rounded corners,
        sharp corners=north,
        boxrule=0.5pt,
        drop shadow=black!50!white,
        attach boxed title to top left={
            xshift=-2mm,
            yshift=-2mm
        },
        boxed title style={
            rounded corners,
            size=small,
            colback=gray!20
        }
    },
    replystyleg/.style={
        enhanced,
        colback=green!15,
        colframe=black,
        colbacktitle=green!30,
        coltitle=black,
        boxrule=0.5pt,
        drop shadow=black!50!white,
        rounded corners,
        sharp corners=north,
        attach boxed title to top right={
            xshift=-2mm,
            yshift=-2mm
        },
        boxed title style={
            rounded corners,
            size=small,
            colback=green!40
        }
    },
    replystyler/.style={
        enhanced,
        colback=red!15,
        colframe=black,
        colbacktitle=red!40,
        coltitle=black,
        boxrule=0.5pt,
        drop shadow=black!50!white,
        rounded corners,
        sharp corners=north,
        attach boxed title to top right={
            xshift=-2mm,
            yshift=-2mm
        },
        boxed title style={
            rounded corners,
            size=small,
            colback=red!40
        }
    }
}

\newtcolorbox{userquery}[1][]{
    userstyle,
    title=Prompt,
    #1
}

\title{Structured Outputs Enable General-Purpose LLMs to be Medical Experts}

\begin{document}
\maketitle
\begin{abstract}
Medical question-answering (QA) is a critical task for evaluating how effectively large language models (LLMs) encode clinical knowledge and assessing their potential applications in medicine. Despite showing promise on multiple-choice tests, LLMs frequently struggle with open-ended medical questions, producing responses with dangerous hallucinations or lacking comprehensive coverage of critical aspects. Existing approaches attempt to address these challenges through domain-specific fine-tuning, but this proves resource-intensive and difficult to scale across models. To improve the comprehensiveness and factuality of medical responses, we propose a novel approach utilizing structured medical reasoning. Our method guides LLMs through an seven-step cognitive process inspired by clinical diagnosis, enabling more accurate and complete answers without additional training. Experiments on the MedLFQA benchmark demonstrate that our approach achieves the highest Factuality Score of 85.8, surpassing fine-tuned models. Notably, this improvement transfers to smaller models, highlighting the method's efficiency and scalability. Our code and datasets are available.


\end{abstract}

\section{Introduction}\label{sec:int}

Language lies at the heart of healthcare delivery, serving as the primary medium through which medical knowledge is communicated and healthcare decisions are made.  Recent advances in LLMs have shown promising potential in medical applications, particularly in tasks requiring deep clinical knowledge and reasoning \cite{singhal2022largelanguagemodelsencode,tu2023generalistbiomedicalai,labrak2024biomistralcollectionopensourcepretrained,jeong2024improvingmedicalreasoningretrieval, liu2023evaluating}. Among these applications, medical QA stands out as a critical task for evaluating how effectively LLMs encode and apply clinical knowledge. This task is particularly challenging, as it requires an understanding of medical context, the recall of relevant knowledge, and the ability to reason with expert-level information.

While existing LLMs have demonstrated impressive performance on medical benchmarks, such as MedMCQA \cite{pal2022medmcqalargescalemultisubject} and USMLE \cite{usmle2023examination}, these evaluations primarily focus on knowledge retrieval rather than comprehensive reasoning. This limitation has led to the development of more challenging benchmarks like MedLFQA \cite{jeong2024olaphimprovingfactualitybiomedical}, which demands detailed, factually accurate, and clinically relevant long-form responses.

\setlength{\textfloatsep}{4pt} 
\setlength{\abovecaptionskip}{4pt} 
\setlength{\belowcaptionskip}{4pt} 

\begin{figure*}
	\centering
	\includegraphics[width=0.9\linewidth]{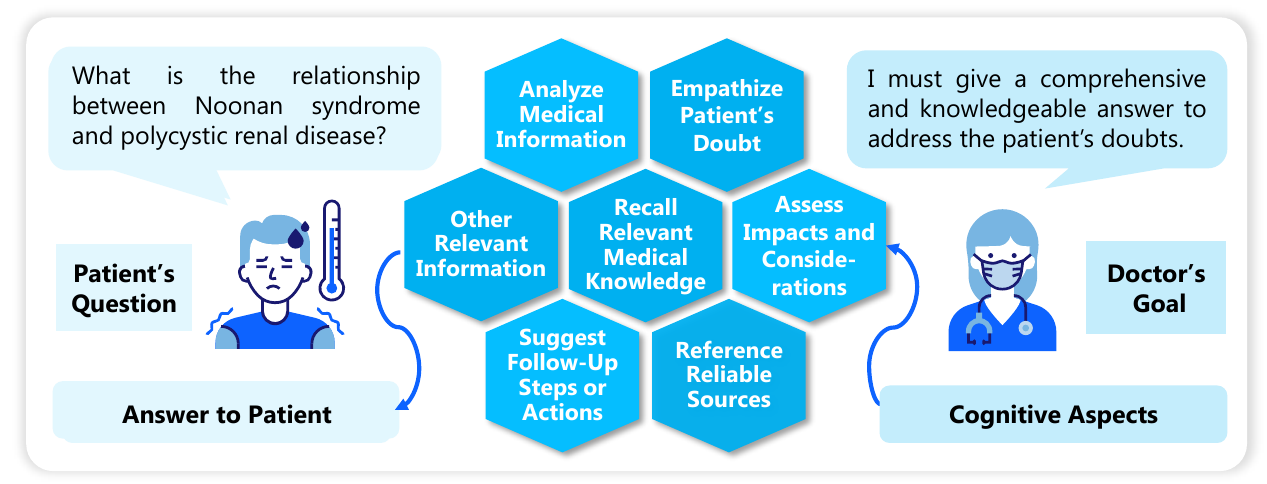}
	\caption{
	A flowchart showing the doctor’s cognitive process to answer a patient’s question, involving medical analysis, relevant information, and follow-up steps. 
		\label{fig:Doctor}}
        \vspace{-6mm} 
\end{figure*}

However, when tasked with generating long-form medical answers, LLMs frequently struggle with two critical challenges. First, they often produce responses containing dangerous hallucinations or factual errors. As shown in Figure~\ref{fig:reason-olaph}, consider a patient's inquiry regarding the concurrent use of Zyrtec and Hydroxyzine:  while an accurate response should detail potential adverse interactions between these antihistamines and emphasize the importance of professional medical guidance, LLMs sometimes generate misleading or incorrect information about drug compatibility. Such inaccuracies could lead to serious clinical consequences if relied upon for medical decision-making.

\begin{figure*}[!tb]
	\centering
	\includegraphics[width=0.9\linewidth]{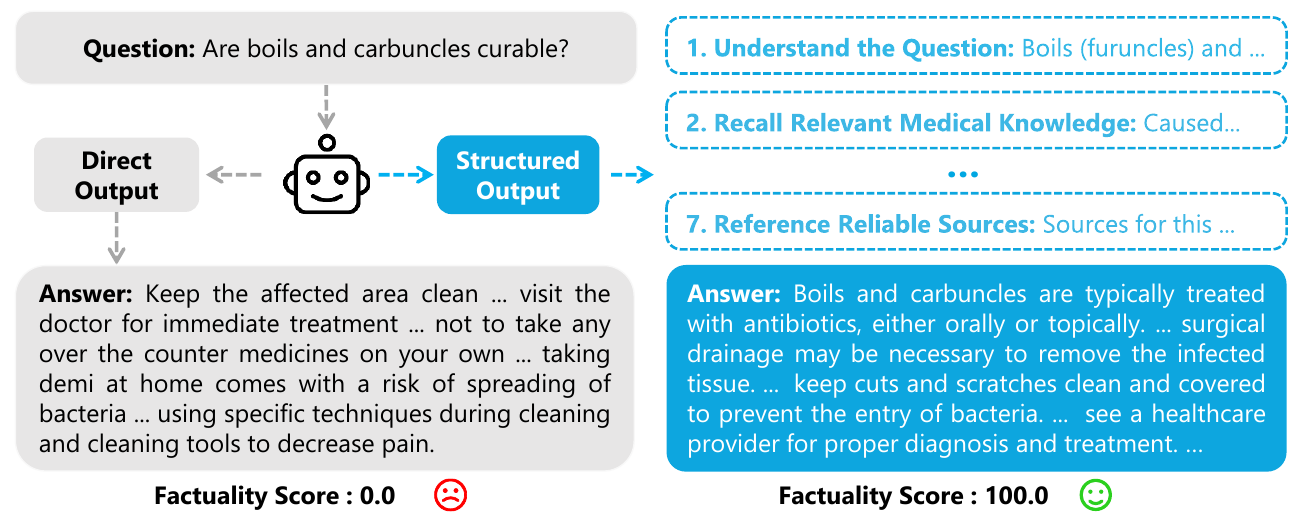}
	\caption{
    Model: Gemma2-7B. LLMs often generate unreliable answers, due to cognitive limitations. Structured output helps LLMs analyze problems step by step, leading to more complete and accurate answers.
    \vspace{-6mm}
\label{fig:compare}}
\end{figure*}

As shown in the Figure~\ref{fig:compare}, the second challenge lies in generating comprehensively informed responses. When asked about a health condition like boils and carbuncles, an adequate response must cover multiple aspects: how they are treated, potential causes, and common symptoms. Current models sometimes provide incomplete responses, addressing only certain parts of the question while leaving out critical information. This lack of comprehensive detail can result in patients or healthcare providers missing important aspects of diagnosis and treatment.

Existing approaches have primarily relied on two strategies to address these challenges: supervised fine-tuning with domain-specific datasets and chain-of-thought (CoT) prompting \cite{kalajdzievski2024scalinglawsforgettingfinetuning}. While fine-tuning can improve performance, it requires substantial computational resources and risks catastrophic forgetting of pre-trained knowledge. CoT prompting has shown promise in enhancing reasoning capabilities, but its effectiveness in generating comprehensive medical responses remains limited, particularly for models with restricted context lengths.

\setlength{\textfloatsep}{4pt} 
\setlength{\abovecaptionskip}{4pt} 
\setlength{\belowcaptionskip}{0pt} 

\begin{figure}[t]
    \centering
    \includegraphics[width=1\linewidth]{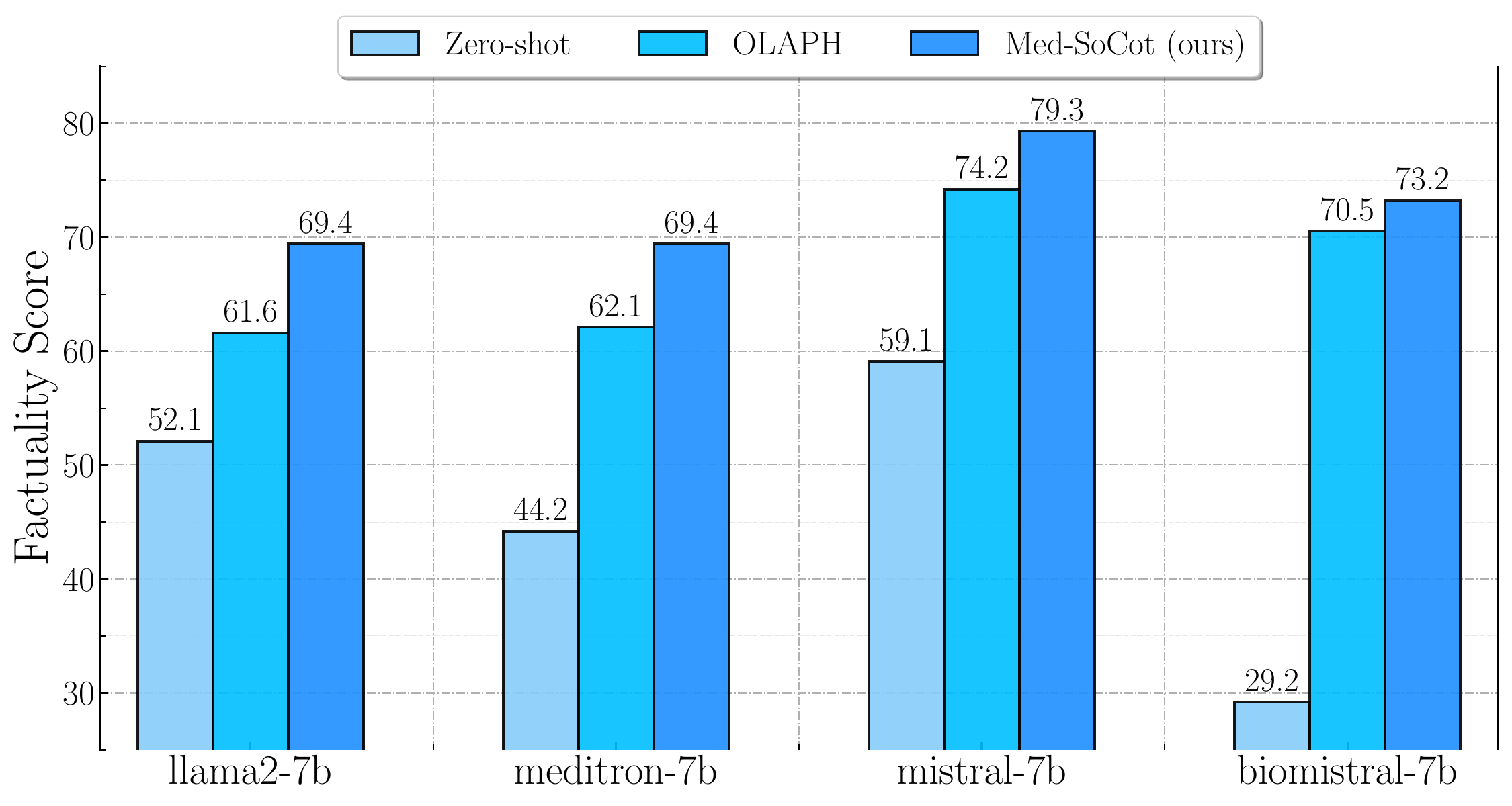}
    \caption{
    Factuality Scores for different models (LLaMA2-7B, Meditron-7B, Mistral-7B, BioMistral-7B) across three methods: Zero-shot, OLAPH, and Med-SoCoT (Ours). 
    }
    \vspace{-4mm}
    \label{fig:cost}
\end{figure}

\setlength{\textfloatsep}{4pt} 
\setlength{\abovecaptionskip}{0pt} 
\setlength{\belowcaptionskip}{10pt} 

\begin{figure}[!tb]
	\centering
	\includegraphics[width=1\linewidth]{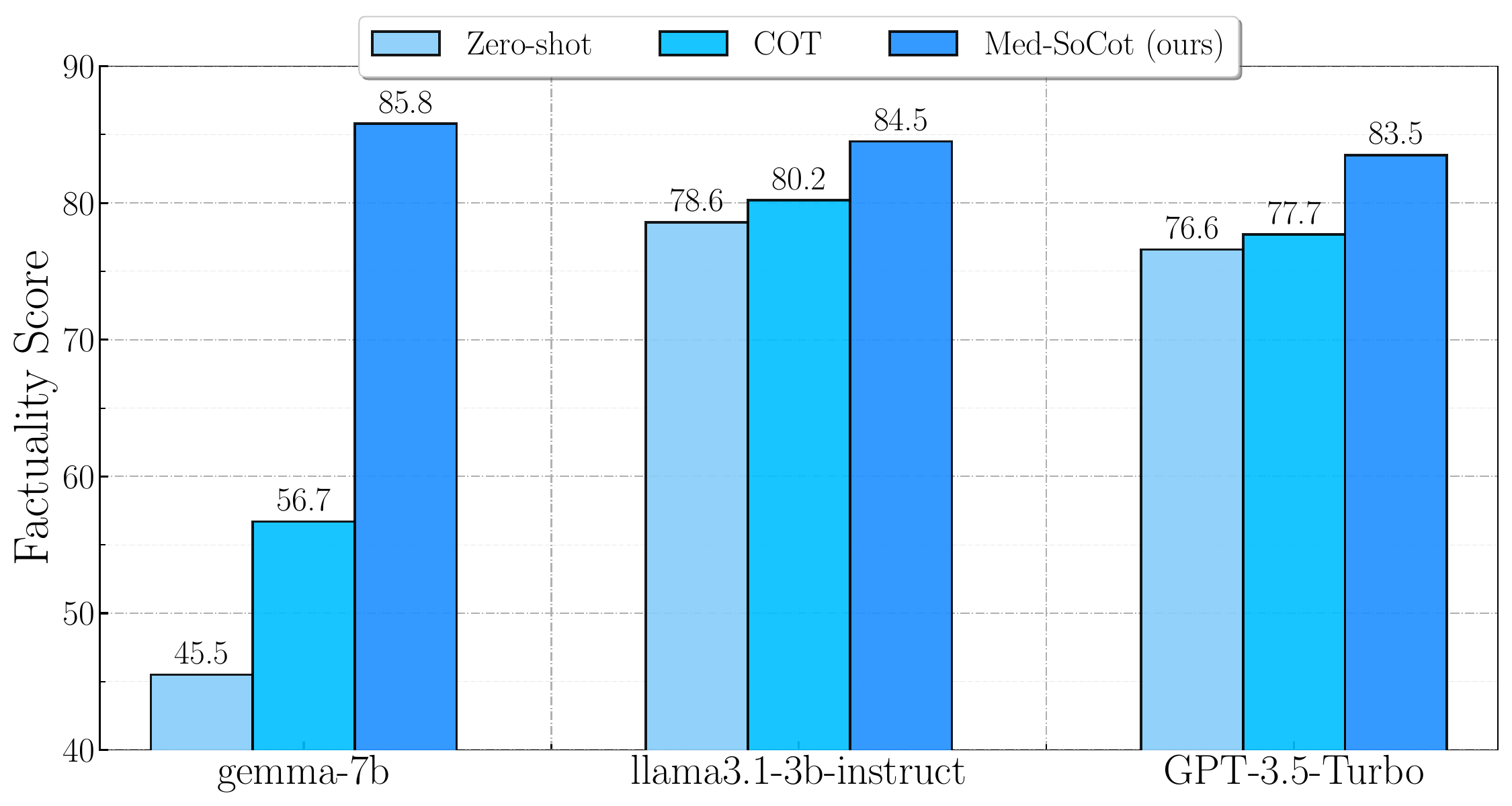}
	\caption{
	Factuality Scores for different models (Gemma-7B, LLaMA3.1-3B-INSTRUCT, GPT-3.5-Turbo) across three methods: Zero-shot, CoT and Med-SoCoT (Ours). 
		\label{fig:2}}
        \vspace{-4mm}
\end{figure}

Drawing inspiration from cognitive science research on how humans process complex information \cite{thagard1996mind, breedlove2010biological, gazzaniga2018cognitive, krathwohl2002revision, duncan2013structure}, we propose a novel medical structured output approach. As shown in the Figure~\ref{fig:Doctor}, this method guides LLMs through a systematic process of medical reasoning, breaking down the response generation into distinct cognitive steps: understanding the question, recalling relevant medical knowledge, analyzing implications, and synthesizing comprehensive answers. By explicitly structuring the output process, we enable LLMs to better organize and validate their medical knowledge, leading to more accurate and complete responses \cite{zhao2024surveylargelanguagemodels}.

\setlength{\textfloatsep}{5pt} 
\setlength{\abovecaptionskip}{2pt} 
\setlength{\belowcaptionskip}{0pt} 

Our research demonstrates that~\textbf{Medical Structured Output CoT (Med-SoCoT)} effectively mitigates hallucinations and "echo" phenomena while improving the factuality in a training-free manner, enabling even smaller models to generate high-quality medical responses. For models with limited context length, overly long prompts can hinder the complete output of structured content. To address this, we adopt a step-by-step reasoning approach, where the model generates one piece of structured content at a time, and the outputs are concatenated for summarization. Notably, this approach notably achieves performance improvements on par with fine-tuned models, demonstrating that prompt engineering alone can substantially enhance factuality without requiring additional training.

Our research focuses on two key questions: First, whether structured output can enhance the factual accuracy of medical question answering; and second, whether smaller models can achieve structured outputs solely through prompt engineering. Our contributions are summarized as follows:
\begin{itemize}
    \setlength\itemsep{-0.5em}
    \item We integrate chain-of-thought reasoning with structured output to develop novel medical prompts, demonstrating that general-purpose LLMs can produce comprehensive answers with reduced hallucinations without additional training. Our approach achieves the highest Factuality Score of 85.8, surpassing even fine-tuned models, which achieve a maximum Factuality Score of 74.2.
    \item For models with shorter context lengths, we employed stepwise reasoning without altering the prompt content, achieving an 12.7\% improvement in the comprehensiveness of the model's output, increasing Factuality Score from 61.6 to 69.4.
    \item For smaller models (3B), our approach remains highly effective, boosting Factuality Score from 78.6 in Zero-shot settings to 84.5. This improvement highlighting the robustness of our method across model sizes.
\end{itemize}

\section{Related Work}

\subsection{Medical question answering}

The landscape of medical question answering (QA) is diverse, encompassing various formats that cater to different aspects of clinical reasoning and knowledge application. These formats include multiple-choice QA \cite{jin2020diseasedoespatienthave, hendrycks2021measuringmassivemultitasklanguage, jin2019pubmedqadatasetbiomedicalresearch, pal2022medmcqalargescalemultisubject, xiong2024benchmarkingretrievalaugmentedgenerationmedicine}, fragment-based QA \cite{krithara2023bioasqqa}, open-domain QA \cite{Lee_2019}, factoid QA \cite{Jin_2022}, multi-turn dialogue QA \cite{bao2023discmedllmbridginggenerallarge}, knowledge-base QA \cite{lozano2023clinfoaiopensourceretrievalaugmentedlarge}, and clinical decision-making QA \cite{hager2024evaluation}. Each format presents unique challenges, ranging from simple knowledge retrieval to complex reasoning and multi-step decision-making, reflecting the multifaceted nature of real-world medical problem-solving.

Among these, Long-Form Question Answering (LFQA) is particularly crucial in the medical field, as it allows for detailed and comprehensive responses to open-ended questions. This is especially important in biomedical and clinical settings, where accuracy and depth are vital for patient care \cite{fan2019eli5longformquestion, stelmakh2023asqafactoidquestionsmeet}. To address these needs, Minbyul developed the MedLFQA dataset by combining several existing QA datasets \cite{abacha2019bridging}. MedLFQA provides a valuable resource for evaluating how large language models (LLMs) handle patient queries, offering insights into the quality of their long-form responses. The dataset includes five distinct medical QA datasets and encompasses various fields like question, answer, and essential details, enabling a thorough assessment of the comprehensiveness and accuracy of LLM-generated medical content.

\subsection{Evaluation of Long-Text Generation}

To evaluate the quality of long text generation by LLMs, ROUGE metrics, such as ROUGE-1, ROUGE-2, and ROUGE-L, are commonly used to assess text similarity at different granularities. To provide a more comprehensive evaluation, we use the Words Composition metric from MedLFQA, which is calculated as shown in Equation~\ref{eq:Words Composition}. This metric averages the F1 scores of ROUGE-1, ROUGE-2, and ROUGE-L, offering a more accurate assessment of the text's quality in terms of word-level, bigram-level, and sentence structure performance.

Some metrics, such as Hallucination Score  and Comprehensiveness Score  \cite{manes2024kqarealworldmedicalqa} , are used to assess the inclusion of key statements in the clinical domain. Specifically, Hallucination Score is a metric designed to measure how many clinical statements in the language model's response contradict factual information. This score is calculated using Equation~\ref{eq:Hal_Score}. Where \( S \) represents all the statements that must be included (Must Have, MH) and those that are optional (Nice to Have, NH), such that \( |S| = |\text{MH}| + |\text{NH}| \). Additionally, Comprehensiveness Score is a metric used to measure how many key clinical statements are included in the language model's response. This score is calculated using Equation~\ref{eq:Com_Score}. To predict the entailment of the response, we used a classification model based on BioBERT \cite{Lee_2019}, which was trained on NLI datasets \cite{bowman2015largeannotatedcorpuslearning, williams2018broadcoveragechallengecorpussentence}.

The Factuality Score is calculated using Equation~\ref{eq:Fac_Score}. Our work is based on these fine-grained and cost-effective evaluation metrics to understand the factuality and lexical structure of LLMs when generating long text.

\subsection{Structured Output}

In traditional NLP literature, structured data typically refers to the extraction of entity-relation-entity triples and attributes from plain text, followed by the construction of knowledge graphs or databases with specific data formats or schemas \cite{jiang2023structgptgeneralframeworklarge, li2023resdsqldecouplingschemalinking}. In previous work, "structuring" was used to reorganize input sentences into a three-layer structure based on the internal semantic relationships between them. This method captures the dependencies and connections among elements in long-text inputs, thereby enhancing LLMs' understanding of knowledge structure and relationships. Earlier studies have explored how structuring input information can improve LLMs' cognitive abilities \cite{liu2024enhancingllmscognitionstructurization}. In the work on GPT-4o, structured output refers to organizing the content generated by a language model into a specific format or pattern \cite{openai2023structuredoutputs}. This output often includes predefined labels, fields, or sections to make it more convenient for users to interpret and use. Structured output not only helps reduce redundancy in generated content but also ensures that the text meets certain requirements or standards.

In this paper, instead of focusing on entity-level information extraction or restructuring the input content, we primarily leverage the model’s structured output capabilities to guide it in completing structured thinking. This enhances the model’s ability to recognize the structure and relationships between recalled knowledge and the posed questions.

\section{Methodology}

\begin{figure*}
	\centering
	\includegraphics[width=1\linewidth]{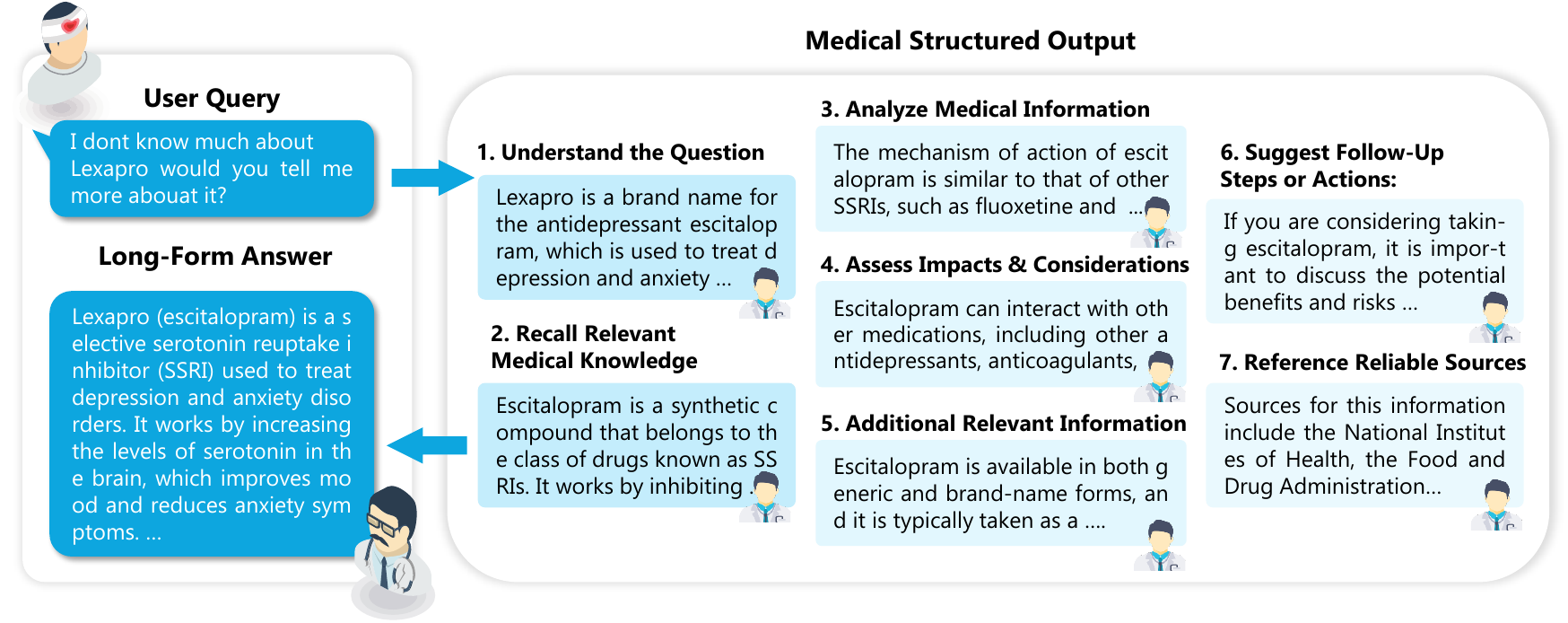}
	\caption{Model: Gemma-7B, Dataset: LiveQA.
    An overview of structured medical reasoning process shows a step-by-step framework to improve the comprehensiveness and factuality of medical QA.
    \vspace{-6mm}
\label{fig:all}}
\end{figure*}

\subsection{Overview}

Structured reasoning plays a crucial role in modern medical applications by enhancing diagnostic accuracy and decision-making efficiency through the systematic organization of complex medical knowledge. By emulating human cognitive processes, it ensures a logical and coherent approach to clinical problem-solving. This is particularly vital for large language models (LLMs) in medical decision-making, where, despite vast knowledge, LLMs often produce fragmented outputs without guidance. A structured framework allows these models to analyze data systematically, consider relevant clinical factors, and generate contextually appropriate, reliable conclusions, improving both accuracy and consistency.

To implement this, we propose a framework based on established clinical reasoning theories \cite{Higgs2020, Young2020, Shin2019, Pelaccia2011}. It breaks down diagnostic and therapeutic reasoning into systematic steps, ensuring a logical progression from patient assessment to treatment planning. Supported by evidence-based strategies and structured models, this approach yields precise, actionable outputs that enhance clinical decision-making.

\subsection{Medical Structured Reasoning Process}

Structured reasoning is a cognitive approach that organizes complex medical problems into systematic decision-making steps. Inspired by clinical reasoning in medicine \cite{Higgs2020, Young2020, Shin2019, Pelaccia2011}, we propose a structured framework comprising seven key cognitive processes. This framework provides large language models (LLMs) with a clear reasoning pathway that mimics human medical cognition, thereby improving accuracy, interpretability, and reducing reasoning errors.

\textbf{1. Understand the Question}

The first step in medical reasoning is accurately understanding the patient’s problem. This involves an in-depth analysis of the chief complaint, medical history, and contextual factors. Research indicates that a well-defined problem statement is the foundation of a successful diagnosis \cite{10.1093/postmj/qgae079}. For LLMs, precise input interpretation is critical in minimizing errors and setting the context for subsequent logical analysis.

\textbf{2. Recall Relevant Medical Knowledge}

Once the problem is clearly defined, clinicians recall relevant disease patterns, risk factors, and physiological mechanisms. The disease script theory suggests that experienced physicians rely on stored patterns for rapid decision-making \cite{Thampy2019}. This step is essential for LLMs, as an effective knowledge retrieval mechanism can reduce hallucinations and ensure medically accurate responses. Moreover, robust knowledge recall directly supports the next step—logical analysis of the medical information.

\textbf{3. Analyze Medical Information}

Using the retrieved knowledge, clinicians systematically compare and interpret the collected patient data to generate differential diagnoses. The dual-process theory highlights that analytical reasoning is particularly crucial in complex or unfamiliar cases \cite{Higgs2020}. For LLMs, this step requires strong logical reasoning capabilities to avoid misinterpretations caused by data noise or biases. Furthermore, accurate knowledge recall (Step 2) enhances this analytical process by providing the necessary context for reasoning.

\textbf{4. Assess Impacts and Considerations}

After forming diagnostic hypotheses, clinicians evaluate potential treatment risks, ethical considerations, and resource constraints. Multilevel decision frameworks, such as the Four Quadrant Model, help balance medical, ethical, and logistical factors \cite{10.3389/fmed.2022.900543}. This step bridges logical analysis (Step 3) and actionable decision-making (Step 5). For LLMs, incorporating structured assessment criteria reduces bias and ensures recommendations align with clinical best practices.

\textbf{5. Provide Additional Relevant Information}

During diagnosis and treatment, clinicians must ensure that all necessary medical information is conveyed clearly to patients and colleagues. Studies show that effective patient education fosters trust and improves treatment adherence \cite{Richards2020}. This step is crucial for LLMs, as providing clear, accessible explanations enhances transparency and interpretability. Additionally, structured information delivery reinforces the justification for proposed actions (Step 6).

\textbf{6. Suggest Follow-Up Steps or Actions}

A well-structured medical decision should include a personalized follow-up plan, specifying additional tests, treatment goals, and monitoring schedules. Research indicates that structured follow-up plans help prevent complications and improve patient adherence \cite{Bae2023}. This step naturally follows information provision (Step 5), ensuring continuity of care. For LLMs, integrating clinical pathways into their reasoning process enables them to generate practical and clinically relevant recommendations.

\textbf{7. Reference Reliable Sources}

Finally, clinical decisions must be grounded in peer-reviewed literature and evidence-based guidelines (e.g., Cochrane Database, PubMed). Medical education research emphasizes that reliance on high-quality information sources is critical for maintaining diagnostic and treatment consistency \cite{Young2020, Pelaccia2011}. This final step strengthens the credibility of previous recommendations and mitigates misinformation risks in LLM-generated medical reasoning.

\textbf{8. Conclusion}

This structured framework in medical decision-making ensures a cohesive and logical process by following a sequential reasoning path, where each step builds upon the previous one. It begins with problem identification, integrates relevant medical knowledge, and systematically analyzes information to enable evidence-based decisions. By assessing potential impacts and ethical considerations, it refines reasoning to ensure contextually appropriate choices. Additionally, it provides supplementary information and follow-up recommendations, supporting both immediate and long-term care strategies. 

Unlike unconstrained generative methods, structured outputs follow predefined frameworks, enhancing consistency, accuracy, and interpretability while reducing errors from fragmented recall or misinterpretation. This approach improves clinical reasoning by guiding models through a step-by-step diagnostic process, minimizing critical oversights. It also strengthens explainability and verifiability, ensuring AI recommendations are transparent, traceable, and aligned with medical guidelines, reinforcing LLM reliability.

\subsection{Structured Output Generation}

\paragraph{Instructional Guidance} To ensure the generated content is logical and coherent, we use prompts or guidewords to direct the model step by step in producing content in a specific format. As shown in Figure~\ref{fig: system_prompt_1}, by clearly specifying the required sections or types of information, the model sequentially fills in the corresponding information, ensuring that the generated content adheres to the expected structure. Additionally, by imposing constraints on the output, the model avoids repetition and redundancy, making the content more concise and intuitive.

\paragraph{Template Generation} By predefining structured templates, the model can populate each field of the template step by step. To enhance the comprehensiveness and accuracy of medical question answering, we designed seven key stages of prompts to guide the model in generating structured content.

\paragraph{Stepwise Generation}Given the complexity of certain tasks that may exceed the model's context processing capabilities. For example, LLaMA2-7B-HF supports a context length of only 4k tokens, we divide them into multiple stages, with the model outputting structured content one step at a time. 
Each generation focuses on a specific stage, and through multiple iterations, the model progressively constructs the complete structured content. After generating the content for all stages, the model combine the information to generate a comprehensive long-form response.

\subsection{Optimization Strategies}

To further enhance the comprehensiveness and credibility of medical question answering, we propose the following optimization strategies:

\paragraph{One-shot Example} After the chain of thought, provide an example closely related to the question to help validate the reasoning process. This example should directly guide towards the final answer, ensuring the accuracy and operational effectiveness of the reasoning chain.
\paragraph{Conciseness of the Problem Statement} Ensure the problem is presented immediately after the chain of thought, avoiding unnecessary spaces, line breaks, or redundant information to improve input clarity and processing efficiency. Refining the problem statement helps reduce potential misinterpretations by the model and improves response time.
\paragraph{Instruction Reinforcement} For critical instructions, repeat them multiple times in the prompt to reinforce their execution. For example, both in the chain of thought and the one-shot example, the "Structured Outputs" tag should be used, and it should be appended again at the end to clearly guide the model's output.
\paragraph{Specialized Markers} Use specific markers to extract important content, such as "ANSWER END," to facilitate subsequent processing and analysis.
\paragraph{Content and Word Count Limitation} We perform quality checks on each step of the step-by-step reasoning, including content and format checks to remove information that is irrelevant to the question in the model output, so that the intermediate reasoning steps do not affect the quality of the answer in the final step. Ensure that the chain of thought is free from redundancy, with each phase of the reasoning chain kept to a maximum of 200 words to avoid repetitive outputs. The final answer should be limited to 512 tokens, ensuring it is concise and precise.
\paragraph{Content Output Strategy} In the first stage, the reasoning chain may output longer structured information (up to 4096 tokens). However, in the second stage, the answer extraction should be limited to 512 tokens to ensure efficiency and focus.

Through the above methods and strategies, we ensure that the structured output for medical question answering is comprehensive, accurate, and meets the high standards required for clinical and research applications.

\section{Experiments}

We evaluated the effectiveness of Medical Structured Outputs CoT (Med-SoCoT) in medical question answering based on general-purpose LLMs, comparing it with the strong baseline method OLAPH \cite{jeong2024olaphimprovingfactualitybiomedical}, which was fine-tuned, whereas our approach was training-free. Our experimental setup closely follows previous work to ensure a fair comparison. Additionally, we validated the effectiveness of our method on other open-source models.

\subsection{Experimental Setup}

We used the MedLFQA benchmark \cite{jeong2024olaphimprovingfactualitybiomedical} for evaluation and adhered to existing work by using factuality as the primary evaluation metric. To facilitate reproducibility, we used five datasets in the benchmark: LiveQA, MedicationQA, HealthSearchQA, K-QA Golden, and K-QA Silver.

\paragraph{Comparison Method} We compared our approach with the state-of-the-art (SoTA) OLAPH method, which utilizes automated evaluation to generate synthetic preference sets and employs preference optimization through iterative learning.

\paragraph{Model Configuration} We applied the structured output method to several popular LLMs, including LLaMA2-7B-HF \cite{touvron2023llama2openfoundation}, Mistral-7B \cite{jiang2023mistral7B}, BioMistral-7B \cite{labrak2024biomistralcollectionopensourcepretrained}, Gemma-7B \cite{gemmateam2024gemmaopenmodelsbased}, GPT-3.5-Turbo. By default, we set the decoding temperature to 0.0 to minimize the randomness in the output of the LLMs. We applied the step-by-step reasoning approach to the LLaMA2-7B-HF, Mistral-7B, BioMistral-7B, and Meditron-7B models, and the direct output approach to the Gemma-7B, LLaMA3.1-3B-INSTRUCT, and GPT-3.5-Turbo models, and reported the Factuality Score.

\subsection{Overall Performance}

\begin{table*}[ht]
\centering

\resizebox{\textwidth}{!}{%
\begin{tabular}{lcccccccccccc}
\toprule
\multirow{2}{*}{\textbf{Method}} & \multicolumn{2}{c}{\textbf{LiveQA}} & \multicolumn{2}{c}{\textbf{MedicationQA}} & \multicolumn{2}{c}{\textbf{HealthSearchQA}} & \multicolumn{2}{c}{\textbf{K-QA Golden}} & \multicolumn{2}{c}{\textbf{K-QA Silver}} & \multicolumn{2}{c}{\textbf{Average}} \\
\cmidrule(lr){2-3} \cmidrule(lr){4-5} \cmidrule(lr){6-7} \cmidrule(lr){8-9} \cmidrule(lr){10-11} \cmidrule(lr){12-13}
   & \textbf{Words} & \textbf{Fact.} & \textbf{Words} & \textbf{Fact.} & \textbf{Words} & \textbf{Fact.} & \textbf{Words} & \textbf{Fact.} & \textbf{Words} & \textbf{Fact.} & \textbf{Words} & \textbf{Fact.} \\
\midrule
LLaMA2-7B w/ Zero-shot & 7.4  & 58.1 & 4.4  & 48.8 & 11.0 & 62.4 & 6.9  & 50.4 & 6.1  & 40.7 & \textcolor[RGB]{30,60,255}{7.2}  & \textcolor[RGB]{30,60,255}{52.1} \\
LLaMA2-7B w/ OLAPH     & 7.7  & 71.2 & 5.1  & 57.1 & 12.0 & 68.2 & 8.4  & 64.3 & 7.5  & 47.4 & \textcolor[RGB]{30,60,255}{8.1}  & \textcolor[RGB]{30,60,255}{61.6} \\
LLaMA2-7B w/ Med-SoCoT (Ours)      & \textbf{7.8}  & \textbf{76.9} & \textbf{7.1}  & \textbf{65.0} & \textbf{13.2} & \textbf{75.1} & \textbf{10.5} & \textbf{72.5} & \textbf{12.2} & \textbf{57.3} & \textcolor[RGB]{30,60,255}{\textbf{10.2}} & \textcolor[RGB]{30,60,255}{\textbf{69.4}} \\
\midrule
Mistral-7B w/ Zero-shot& 8.5  & 59.6 & 5.4  & 56.6 & 15.8 & 78.7 & 9.8  & 57.9 & 8.4  & 42.8 & \textcolor[RGB]{30,60,255}{9.6}  & \textcolor[RGB]{30,60,255}{59.1} \\
Mistral-7B w/ OLAPH    & 6.6  & 66.8 & 6.3  & 67.5 & 13.7 & 83.8 & 10.9 & 75.2 & \textbf{18.2} & 77.5 & \textcolor[RGB]{30,60,255}{11.1} & \textcolor[RGB]{30,60,255}{74.2} \\
Mistral-7B w/ Med-SoCoT (Ours)     & \textbf{9.6}  & \textbf{80.6} & \textbf{8.2}  & \textbf{70.7} & \textbf{14.5} & \textbf{85.8}& \textbf{12.1} & \textbf{80.8} & 12.2 & \textbf{78.8} & \textcolor[RGB]{30,60,255}{\textbf{11.3}} &  \textcolor[RGB]{30,60,255}{\textbf{79.3}}\\
\midrule
BioMistral-7B$^\clubsuit$ w/ Zero-shot & 4.7 & 27.4 & 2.1 & 12.9 & 7.0 & 41.1 & 7.5 & 37.0 & 5.4 & 27.5 & \textcolor[RGB]{30,60,255}{5.3}  & \textcolor[RGB]{30,60,255}{29.2} \\
BioMistral-7B$^\clubsuit$ w/ OLAPH & 13.5 & 69.1 & 12.5 & 70.9 & 18.4 & 76.9 & 17.3 & 75.6 & 17.2 & 59.8 & \textcolor[RGB]{30,60,255}{15.8} & \textcolor[RGB]{30,60,255}{70.5} \\
BioMistral-7B$^\clubsuit$ w/ Med-SoCoT (Ours)  & \textbf{14.9}  & \textbf{71.1} & \textbf{15.5}  & \textbf{73.6} & \textbf{20.5} & \textbf{78.7} & \textbf{17.5} & \textbf{79.4} & \textbf{17.8} & \textbf{63.0} & \textcolor[RGB]{30,60,255}{\textbf{17.3}} & \textcolor[RGB]{30,60,255}{\textbf{73.2}} \\
\midrule
Meditron-7B$^\clubsuit$ w/ Zero-shot & 6.5 & 49.5 & 3.7 & 44.0 & 7.4 & 45.7 & 6.0 & 44.6 & 5.5 & 37.3 & \textcolor[RGB]{30,60,255}{5.8}  & \textcolor[RGB]{30,60,255}{44.2} \\
Meditron-7B$^\clubsuit$ w/ OLAPH   & 8.0  & 66.7 & 5.9  & 62.7 & 8.7  & 50.4 & 10.4 & 71.3 & 11.0 & 59.6 & \textcolor[RGB]{30,60,255}{8.8}  & \textcolor[RGB]{30,60,255}{62.1} \\
Meditron-7B$^\clubsuit$ w/ Med-SoCoT (Ours)    & \textbf{10.1}  & \textbf{69.9} & \textbf{6.4}  & \textbf{63.6} & \textbf{9.3}  & \textbf{70.3} & \textbf{12.5}  & \textbf{72.2} & \textbf{13.4}  & \textbf{71.2} & \textcolor[RGB]{30,60,255}{\textbf{10.3}} & \textcolor[RGB]{30,60,255}{\textbf{69.4}} \\
\bottomrule
\end{tabular}%
}
\caption{Performance of models (LLaMA2-7B, Mistral-7B, BioMistral-7B, Meditron-7B) on MedLFQA datasets, evaluated on Words Composition (Words) and Factuality Score (Fact.). Models are tested in Zero-shot (no prompt), fine-tuned with OLAPH, and with our Med-SoCoT method. Models with $^\clubsuit$ are medically fine-tuned. The best result for each model is bolded, and average values are in blue.}
\label{tab:results1}
\end{table*}

This chapter examines whether general and medical models enhance factual scores via structured output and whether models with limited context length benefit from step-by-step reasoning. Additionally, it analyzes the impact of structured output on text quality, including Words Composition indicators. Below is a detailed analysis:

Table~\ref{tab:results1} shows that both general models (LLaMA2-7B, Mistral-7B) and medical fine-tuned models (BioMistral-7B, Meditron-7B) significantly improve Factuality Score using the Med-SoCoT method. This demonstrates the effectiveness of structured output. Notably, Mistral-7B achieves a factual score of 79.3, surpassing other methods.

Table~\ref{tab:results1} shows that for models with limited context length, step-by-step reasoning enhances both structured output and factuality scores. For example, after applying Med-SoCoT, Mistral-7B achieved high factuality scores, particularly on the HealthSearchQA  datasets, with scores of 85.8.

In terms of Words Composition, the Med-SoCoT method outperformed the OLAPH method, showing significant improvements across multiple models. For example, Mistral-7B and BioMistral-7B achieved Words Composition scores of 11.3 and 17.3, respectively, surpassing OLAPH.

\subsection{Additional Experiments}

\begin{table*}[t]
    \centering
    \begin{adjustbox}{width=\linewidth}
        \setlength{\tabcolsep}{3pt}
        \begin{tabular}{@{}l c c c c c c c c c c c c c c c c@{}}
            \toprule
            \textbf{Method} & \multicolumn{2}{c}{\textbf{LiveQA}} & \multicolumn{2}{c}{\textbf{MedicationQA}} & \multicolumn{2}{c}{\textbf{HealthSearchQA}} & \multicolumn{2}{c}{\textbf{K-QA Golden}} & \multicolumn{2}{c}{\textbf{K-QA Silver}} & \multicolumn{2}{c}{\textbf{Average}} \\
            \cmidrule(lr){2-3} \cmidrule(lr){4-5} \cmidrule(lr){6-7} \cmidrule(lr){8-9} \cmidrule(lr){10-11} \cmidrule(lr){12-13}
            & \textbf{Words} & \textbf{Fact.} & \textbf{Words} & \textbf{Fact.} & \textbf{Words} & \textbf{Fact.} & \textbf{Words} & \textbf{Fact.} & \textbf{Words} & \textbf{Fact.} & \textbf{Words} & \textbf{Fact.} \\
            \midrule
            Gemma-7B w/ Zero-shot
            & 8.3 & 55.7 & 6.0 & 41.1 & 12.4 & 58.0 & 12.4 & 40.9 & 6.1 & 31.9 & \color[RGB]{30,60,255}{9.1} & \color[RGB]{30,60,255}{45.5} \\
            Gemma-7B w/ CoT
            & 12.7 & 60.4 & 6.8 & 55.7 & 12.9 & 69.9 & 8.6 & 55.9 & 8.2 & 41.3 & \color[RGB]{30,60,255}{9.8} & \color[RGB]{30,60,255}{56.7} \\
            Gemma-7B w/ Med-SoCoT (Ours)
            & \textbf{12.7} & \textbf{86.6} & \textbf{10.2} & \textbf{83.0} & \textbf{23.3} & \textbf{92.2} & \textbf{17.9} & \textbf{83.8} & \textbf{21.1} & \textbf{83.4} & \color[RGB]{30,60,255}{\textbf{17.0}} & \color[RGB]{30,60,255}{\textbf{85.8}} \\
            \midrule
            LLaMA3.1-3B-INSTRUCT w/ Zero-shot
            & 10.6 & 82.1 & 9.6 & 76.6 & 21.2 & 82.8 & 15.8 & 82.7 & 18.8 & 69.0 & \color[RGB]{30,60,255}{15.2} & \color[RGB]{30,60,255}{78.6} \\
            LLaMA3.1-3B-INSTRUCT w/ CoT
            & 11.0 & 82.5 & 10.7 & 77.6 & 20.1 & 83.7 & 15.3 & 82.6 & 18.9 & 74.7 & \color[RGB]{30,60,255}{15.2} & \color[RGB]{30,60,255}{80.2} \\
            LLaMA3.1-3B-INSTRUCT w/ Med-SoCoT (Ours)
            & \textbf{11.5} & \textbf{86.0} & \textbf{10.9} & \textbf{81.6} & \textbf{21.8} & \textbf{87.6} & \textbf{15.9} & \textbf{84.5} & \textbf{19.2} & \textbf{82.7} & \color[RGB]{30,60,255}{\textbf{15.8}} & \color[RGB]{30,60,255}{\textbf{84.5}} \\
            \midrule
            GPT-3.5-Turbo$^\spadesuit$ w/ Zero-shot
            & 28.8 & 77.7 & 38.2 & 79.2 & 29.7 & 74.0 & 35.6 & 76.3 & 36.2 & 75.7 & \color[RGB]{30,60,255}{33.7} & \color[RGB]{30,60,255}{76.6} \\
            GPT-3.5-Turbo$^\spadesuit$ w/ CoT
            & 29.3 & 78.2 & 40.6 & 80.1 & 30.9 & 76.5 & 37.8 & 76.6 & 38.6 & 77.3 & \color[RGB]{30,60,255}{35.4} & \color[RGB]{30,60,255}{77.7} \\
            GPT-3.5-Turbo$^\spadesuit$ w/ Med-SoCoT (Ours)
            & \textbf{31.2} & \textbf{87.2} & \textbf{43.2} & \textbf{80.8} & \textbf{37.6} & \textbf{78.1} & \textbf{45.2} & \textbf{87.6} & \textbf{45.3} & \textbf{83.7} & \color[RGB]{30,60,255}{\textbf{40.5}} & \color[RGB]{30,60,255}{\textbf{83.5}} \\

            \bottomrule
        \end{tabular}
    \end{adjustbox}
    \caption{Performance of models (Gemma-7B, LLaMA3.1-3B-INSTRUCT, GPT-3.5-Turbo) on MedLFQA datasets, evaluated on Words Composition (Words) and Factuality Score (Fact.). Models are tested in Zero-shot (no prompt), CoT (Chain-of-Thought), and with our Med-SoCoT method. Models with $^\spadesuit$ are closed source. The best result for each model is bolded, and average values are in blue.}

    \label{tab:gemma_llama3_comparison}
    \vspace{-6mm}
    
\end{table*}

According to Table~\ref{tab:gemma_llama3_comparison}, in the additional experiments, we compared the effects of standard Chain-of-Thought (CoT) prompts and structured output prompts, while also evaluating the structured output capabilities of both the 3B small model and general large models. The experimental results show that structured output significantly improves the factuality score while maintaining text quality, for both the open-source Gemma-7B model and the closed-source GPT-3.5-Turbo model. For instance, when Gemma-7B adopted the Med-SoCoT method, its factuality score increased from 56.7 to 85.8, demonstrating a remarkable improvement. Similarly, GPT-3.5-Turbo's factuality score improved from 77.7 to 83.5 with structured output, further confirming the effectiveness of this approach.

Additionally, for the latest 3B small model, LLaMA3.1-3B-INSTRUCT, its strong instruction-following ability and contextual understanding allow it to generate structured content in a single step without requiring iterative reasoning. Experimental results show that after adopting the Med-SoCoT method, LLaMA3.1-3B-INSTRUCT's factuality score increased from 80.2 to 84.5, approaching the performance of GPT-3.5-Turbo. This indicates that the small-scale model has strong potential in structured output tasks.

\subsection{Ablation Study}

Our method employs the Med-SoCoT prompt, and we conducted ablation studies on the MedLFQA dataset using the Gemma-7B model to assess the impact of key steps on performance. The studies focused on the contribution of steps like "Understanding the Question," "Analyzing Medical Information," and "Suggesting Follow-Ups" to the Factuality Score. The results are presented in Tables~\ref{tab:step_importance}, \ref{tab:step_combinations}, and \ref{tab:step_order}.

Table~\ref{tab:step_importance} demonstrates that removing core steps significantly degrades performance, highlighting their role in improving reasoning and coherence. Table~\ref{tab:step_combinations} confirms that retaining core steps leads to better results, while Table~\ref{tab:step_order} shows that altering the order of core steps harms performance, emphasizing the need for a logical sequence. 

Additionally, Table~\ref{tab:abl_con} reveals that removing features like One-shot Examples, Instruction Reinforcement, and Specialized Markers leads to a significant drop in the Factuality Score, highlighting their importance in task understanding and clarity.

In summary, our findings validate that the Med-SoCoT prompt, with its carefully structured steps and technical components, is crucial for optimizing performance and reasoning capabilities. Further experimental data and case studies are available in the Appendix.

\section{Conclusion}

This paper introduces a method centered on structured outputs to improve the factuality and comprehensiveness of LLMs in medical QA tasks. By leveraging structured prompts and supporting both one-shot reasoning and stepwise reasoning, the method enhances long-form text generation without additional training. Structured outputs significantly improve factual accuracy, enabling smaller models to perform comparably to fine-tuned large models, while stepwise reasoning helps mitigate performance limitations in resource-constrained settings. The structured prompts, covering seven key aspects of medical reasoning, consistently outperform baselines in  factuality accuracy and comprehensiveness. This approach demonstrates the potential of structured outputs in optimizing LLM performance across specialized domains, offering a cost-effective alternative to model training with broader applicability in areas like legal and technical documentation.

\section*{Limitations}
Our Med-SoCoT prompt is an initial step toward optimizing structured outputs in medical reasoning tasks. However, its generalizability to non-medical domains and efficiency in resource-constrained settings remain open questions. Future work could explore more advanced prompt designs and adaptive strategies to enhance performance and applicability across diverse tasks and models.

\section*{Ethical Considerations}
Ethical considerations are central to our research. In this study, we ensure adherence to ethical principles by exclusively using publicly available datasets and employing models that are open-source or widely accepted within the research community. We emphasize transparency in all stages of our work and prioritize the responsible application of technology, particularly in the sensitive domain of medical reasoning, to ensure that our contributions promote fairness, reliability, and societal benefit.

\bibliographystyle{acl_natbib}
\bibliography{acl2020}

\newpage

\appendix
\onecolumn
\section{Evaluation Index Calculation Method} \label{sec:eval_index}

In this section, we describe the formulas used to compute various evaluation metrics.

\subsection{Words Composition (Words Composition)}
The Words Composition metric is calculated as the average of ROUGE-1, ROUGE-2, and ROUGE-L F1 scores:
\[
\text{Words Composition} = \frac{\text{ROUGE-1 F1} + \text{ROUGE-2 F1} + \text{ROUGE-L F1}}{3} 
\label{eq:Words Composition}
\]

\subsection{Hallucination Score (Hallucination Score)}
The Hallucination Score measures the proportion of elements in set \(S\) that are contradicted by the prediction \(\hat{P}\):
\[
\text{Hallucination Score}(\hat{P}) = \frac{|\{x \in S \mid \hat{P} \text{ contradicts } x\}|}{|S|}
\label{eq:Hal_Score}
\]

\subsection{Comprehension Score (Comprehensiveness Score)}
The Comprehension Score evaluates the proportion of elements in the medical hypothesis set \(MH\) that are entailed by the prediction \(\hat{P}\):
\[
\text{Comprehensiveness Score}(\hat{P}) = \frac{|\{x \in MH \mid \hat{P} \text{ entails } x\}|}{|MH|}
\label{eq:Com_Score}
\]

\subsection{Factual Score (Factuality Score)}
The Factual Score combines the Comprehension Score and the Hallucination Score to produce a balanced metric, normalized to a scale of 0 to 100:
\[
\text{Factuality Score} = \frac{(\text{Comprehensiveness Score} - \text{Hallucination Score}) + 100}{2}
\label{eq:Fac_Score}
\]

\section{Experiment Settings}

\begin{table*}[ht]
    \centering
    \begin{adjustbox}{width=\linewidth}
        \setlength{\tabcolsep}{4pt}
        \begin{tabular}{@{}llcccccc@{}}
            \toprule
            \multicolumn{1}{c}{\textbf{Dataset}}      & \multicolumn{1}{c}{\textbf{Format}} & \begin{tabular}[c]{@{}c@{}}\textbf{\# of QA} \\ \textbf{pairs}\end{tabular} & \begin{tabular}[c]{@{}c@{}}\textbf{\# of Ambiguous} \\ \textbf{Questions}\end{tabular} & \begin{tabular}[c]{@{}c@{}}\textbf{Avg. Length} \\ \textbf{of Answers}\end{tabular} & \begin{tabular}[c]{@{}c@{}}\textbf{Avg. \# of} \\ \textbf{MH statements}\end{tabular} & \begin{tabular}[c]{@{}c@{}}\textbf{Avg. \# of} \\ \textbf{NH Statements}\end{tabular} \\ \midrule

            LiveQA \cite{abacha2017overview} & (Q, A, MH, NH) & {\large 100} & {\large 4} & {\large 82.8} & {\large 2.6} & {\large 2.3} \\
            MedicationQA \cite{abacha2019bridging} & (Q, A, MH, NH) & {\large 666} & {\large 24} & {\large 55.5} & {\large 2.6} & {\large 2.3} \\
            HealthSearchQA \cite{singhal2023large} & (Q, A, MH, NH) & {\large 3,077} & {\large 96} & {\large 118.8} & {\large 2.6} & {\large 2.3} \\
            K-QA Golden \cite{manes2024kqarealworldmedicalqa} & (Q, A, MH, NH) & {\large 201} & {\large 1} & {\large 88.5} & {\large 4.4} & {\large 3.5} \\
            K-QA Silver \cite{manes2024kqarealworldmedicalqa} & (Q, A, MH, NH) & {\large 904} & {\large 106} & {\large 99.9} & {\large 2.4} & {\large 2.0} \\
            \bottomrule
        \end{tabular}
    \end{adjustbox}
    \caption{Overview of MedLFQA datasets with formats and key statistics.}
    \label{tab:dataset_overview}
    \vspace{0mm}
\end{table*}

\section{Model Settings}

\begin{table}[ht]
    \centering 
    \small 

    \label{tab:model_details}
    \renewcommand{\arraystretch}{1} 
    \begin{tabular}{l l l l}
        \toprule
        \textbf{Model}             & \textbf{Method} & \textbf{Context Length} & \textbf{Type}            \\ 
        \midrule
        LLaMA2-7B-HF               & STEPWISE        & 512                    & General                  \\ 
        Mistral-7B                & STEPWISE        & 512                    & General                  \\ 
        BioMistral-7B             & STEPWISE        & 512                    & Medical specific         \\ 
        Meditron-7B              & STEPWISE        & 512                      & Medical specific         \\
        Gemma-7B                   & DIRECT          & 2048                   & General                  \\ 
        LLaMA3.1-3B-INSTRUCT      & DIRECT          & 4096                   & General                  \\ 
        GPT-3.5-Turbo       & DIRECT          & 4096                   & General                  \\ 
        \bottomrule
    \end{tabular}
    \caption{Specific Model Settings Adopted by the Med-SoCoT Method}
\end{table}

\section{Ablation Study}

\begin{table}[t]
    \centering
    
        \begin{tabular}{@{}lccc@{}}
            \toprule
            \textbf{Gemma-7B} & \textbf{Step Removed} & \textbf{Factuality Score} & \textbf{$\Delta$} \\
            \midrule
            Baseline (7-step Full)      & None                  & 71.6 & - \\
            w/o Step 1                  & Understand the Question & 66.5 & $\downarrow 5.1$ (7.1\%) \\
            w/o Step 2                  & Recall Medical Knowledge & 70.4 & $\downarrow 1.2$ (1.7\%) \\
            w/o Step 3                  & Analyze Medical Info    & 65.0 & $\downarrow 6.6$ (9.2\%) \\
            w/o Step 4                  & Assess Impacts          & 69.8 & $\downarrow 1.8$ (2.5\%) \\
            w/o Step 5                  & Add Relevant Info       & 70.0 & $\downarrow 1.6$ (2.2\%) \\
            w/o Step 6                  & Suggest Follow-Ups      & 66.0 & $\downarrow 5.6$ (7.8\%) \\
            w/o Step 7                  & Reference Sources       & 70.5 & $\downarrow 1.1$ (1.5\%) \\
            \bottomrule
        \end{tabular}
  
    \caption{Ablation Study on the Importance of Each Step in the Med-SoCoT Using the Gemma-7B on the MedLFQA.}
    \label{tab:step_importance}
\end{table}

\begin{table}[t]
    \centering

        \begin{tabular}{@{}lccc@{}}
            \toprule
            \textbf{Gemma-7B } & \textbf{Steps Retained} & \textbf{Factuality Score} & \textbf{$\Delta$} \\
            \midrule
            Baseline (7-step Full)      & All Steps               & 71.6 & - \\
            Step 1 + Step 3 + Step 6    & Core Steps Combination   & 60.5 & $\downarrow 11.1$ (15.5\%) \\
            Step 1 + Step 3             & Only Two Core Steps      & 55.0 & $\downarrow 16.6$ (23.2\%) \\
            Step 2 + Step 4 + Step 5    & Non-Core Steps Combination & 58.7 & $\downarrow 12.9$ (18.0\%) \\
            \bottomrule
        \end{tabular}

    \caption{Ablation Study on the Effectiveness of Different Step Combinations in the Med-SoCoT Using the Gemma-7B Model on the MedLFQA.}
    \label{tab:step_combinations}
\end{table}

\begin{table}[t]
    \centering

        \begin{tabular}{@{}lccc@{}}
            \toprule
            \textbf{Gemma-7B} & \textbf{Step Order Adjusted} & \textbf{Factuality Score} & \textbf{$\Delta$} \\
            \midrule
            Baseline (7-step Full)      & Standard Order              & 71.6 & - \\
            Step 3 $\leftrightarrow$ Step 6 & Swap Step 3 and Step 6   & 70.5 & $\downarrow$ 1.1 (1.5\%) \\
            Step 1 $\leftrightarrow$ Step 4 & Swap Step 1 and Step 4   & 70.8 & $\downarrow$ 0.8 (1.1\%) \\
            Step 5 $\leftrightarrow$ Step 7 & Swap Step 5 and Step 7   & 71.2 & $\downarrow$ 0.4 (0.6\%) \\
            \bottomrule
        \end{tabular}

    \caption{Ablation Study on the Impact of Step Order Adjustments in the Med-SoCoT Using the Gemma-7B Model on the MedLFQA.}
    \label{tab:step_order}
\end{table}

\begin{table}[t]
	\centering

		\begin{tabular}{@{}lccc@{}}
			\toprule
			\textbf{Gemma-7B Prompt Version} & \textbf{Factuality Score} & \textbf{$\Delta$} \\
			\midrule
			\midrule
			Baseline (Med-SoCoT) & 69.4 & - \\
			w/o One-shot Example    & 65.2 & $\downarrow 6.0\%$ \\
			w/o Instruction Reinforcement  & 66.8 & $\downarrow 3.7\%$ \\
			w/o Specialized Markers      & 67.5 & $\downarrow 2.7\%$ \\
			w/o All Features        & 60.3 & $\downarrow 13.1\%$ \\
			\bottomrule
		\end{tabular}
	
	\caption{Ablation Study Results of Gemma-7B on Med-SoCoT Prompt Variations, Evaluating the Impact of One-shot Examples, Instruction Reinforcement, and Specialized Markers on Factuality Score Performance.  
    \vspace{-5mm}}
	\label{tab:abl_con}
\end{table}
\vfill
\pagebreak

\afterpage{\clearpage}
\section{Example Analysis}

\begin{figure}[ht]
    \centering
\begin{tcolorbox}[fonttitle = \small\bfseries, title=Reasoning without prompt,colframe=gray!2!black,colback=gray!2!white,boxrule=1pt,boxsep=2pt,left=5pt,right=5pt,fontupper=\footnotesize, halign title = flush center]
\textbf{Question:}  I take zyrtec for my allergies will hydroxyzine be ok to take with it?
\tcbline
\textbf{\color[RGB]{0,0,0}{Answer}:} Can Zyrtec (cetirizine) be combined with hydroxyzine? Absolutely! In general, cetirizine for the histamine issues and hydroxyzine you like probably for the itching component. Can I mix Benadryl and hydroxyzine? 
\tcbline
\textbf{\color[RGB]{0,0,0}{Factuality Score}:} 27.4 {\color[RGB]{255,0,0}{\xmark}}
\end{tcolorbox}
\begin{tcolorbox}
[fonttitle = \small\bfseries, title=Reasoning with OLAPH,colframe=gray!2!black,colback=gray!2!white,boxrule=1pt,boxsep=2pt,left=5pt,right=5pt,fontupper=\footnotesize, halign title = flush center]
\textbf{Question:}  I take zyrtec for my allergies will hydroxyzine be ok to take with it?
\tcbline
\textbf{\color[RGB]{0,0,0}{Answer}:} Hydroxyzine is a diphenhydramine. It is an antihistamine that reduces the effects of natural chemical histamine in the body. Hydroxyzine is used to treat or prevent the symptoms of an allergic reaction. ... Zyrtec is a brand name for the drug cetirizine. It is an antihistamine that reduces the effects of natural chemical histamine in the body. Zyrtec is used to treat or prevent the symptoms of an allergic reaction. Zyrtec is also used to treat the itching and swelling of a skin rash caused by a condition called urticaria.
\tcbline
\textbf{\color[RGB]{0,0,0}{Factuality Score}:} 58.4 {\color[RGB]{255,0,0}{\xmark}}
\end{tcolorbox}

    \caption{Dataset: K-QA Silver \cite{manes2024kqarealworldmedicalqa}, Model: Mistral-7B \cite{jiang2023mistral7B} (\textit{Upper}) LLM directly answers questions, resulting in a large number of hallucinations
    (\textit{Lower}) 
    After training with the OLAPH framework, although it can answer relevant knowledge, it cannot provide accurate answers.
    }
    \label{fig:reason-olaph}
\end{figure}

\begin{figure}[ht]
    \centering

\begin{tcolorbox}[fonttitle = \small\bfseries, title=Reasoning with Structed Outputs (Ours),colframe=gray!2!black,colback=gray!2!white,boxrule=1pt,boxsep=2pt,left=5pt,right=5pt,fontupper=\footnotesize, halign title = flush center]
\textbf{Question:}  I take zyrtec for my allergies will hydroxyzine be ok to take with it?
\tcbline
\textbf{\color[RGB]{0,0,0}{Structured Output}:}

\textbf{\color[RGB]{0,0,0}{1. Understand the Question:}} 
Zyrtec is an antihistamine used to treat allergic reactions, while Hydroxyzine is also an anti-histamine but used for anxiety and insomnia. Both medications can cause drowsiness, so it is important to be aware of potential interactions and side effects. 

\textbf{\color[RGB]{0,0,0}{2. Recall Relevant Medical Knowledge:}} 
Hydroxyzine is an antihistamine that is used to treat allergic reactions, anxiety, insomnia, nausea, vomiting, ... , watery eyes, cough, wheeziness, asthma, hay fever, insect bites, stings, allergens, ... , infections, viruses, bacteria, fungi, parasites, toxins, poisons, venoms, snake bit, spider bit. 

\textbf{\color[RGB]{0,0,0}{3. Analyze Medical Information:}} 
Zyrtec is a brand name for cetirizine, an antihistamine used for allergies. Zyrtec is also used as a sedative, hypnotic, antidepressant, and anti-anxiety medication. Hydroxyzine and Zyrtec are both antihistamines. They are used in different ways. The main difference between hydroxyzine and cetirizine is that hydroxyzine is more sedating than cetirizine. Cetirizine has less sedation than hydroxyzine. Both hydroxyzine and cetirizine are antihistamines. Antihistamines are drugs that block the effects of histamine. ...
\tcbline
\textbf{\color[RGB]{0,0,0}Answer:} Zyrtec is one of the most popular anticholinergic drugs used today. This drug is often used by people who suffer from allergies. However, this drug can also be used with other drugs. One of these drugs is hydrocortisone.{\color[RGB]{71,162,255}{When you take this medication, you will experience a lot of side-effects. These side effect include dizziness and drowsy. You should not take hydrocodone with Zyrtec. If you are taking this medicine, then you should consult your doctor first. }} 
\tcbline
\textbf{\color[RGB]{0,0,0}{Factuality Score}:} 100.0 {\color[RGB]{0,255,0}{\checkmark}}
\end{tcolorbox}
    \caption{Dataset: K-QA Silver \cite{manes2024kqarealworldmedicalqa}, Model: Mistral-7B \cite{jiang2023mistral7B} Use structured outputs and answer questions that include all  {\color[RGB]{71,162,255}{correct statements}}.
    }
    \label{fig:reason-so}
\end{figure}

\clearpage

\section{Prompt}

\begin{figure*}[ht]
    \begin{userquery}
    \fontsize{10.5}{12}\selectfont

    \textbf{Medical Structured Output Prompt (part1)}: 
    
\# Task: You are a helpful assistant. Step-by-Step Thinking for Structured Medical Question Answering.

\#\# General Instructions:
- Generate detailed and structured medical responses based on the given medical question. Answers should be grounded in current medical knowledge, covering all key aspects of the question.

- Ensure the answer includes background, etiology, symptoms, diagnosis, treatment, and prevention.

- The answer should be logically organized and provide accurate, comprehensive medical information.

\#\# Task Instructions:
- Generate a comprehensive response based on the input question. The response should cover everything from background information to diagnosis and treatment recommendations, ensuring a structured and coherent output.

- The answer should address as many aspects of the medical question as possible, considering risk factors, complications, and related medical conditions.

- Consider the relationship between diseases and medications.

- Do not output duplicate content.

- Each thought process should not exceed 200 words.

\#\# Output Structure:
- The output should follow the structured template below to ensure the completeness and professionalism of the medical response.

- Please ensure that the output contains Long-Form Answer

\end{userquery}
\caption{Medical Structured Output Prompt (part1).}
\label{fig: system_prompt_1}
\end{figure*}

\begin{figure*}[ht]
    \begin{userquery}
    \fontsize{10.5}{12}\selectfont

    \textbf{Medical Structured Output Prompt  (part2)}: 

\#\# Chain of Thought:

\#\#\# 1. Understand the Question:

- {{Explain the background or definition of the medical issue. Provide a brief description of basic concepts and possibly affected systems or organs.}}

- {{Identify and define key medical terms and concepts.}}

- {{Clarify the specific information or details requested.}}

\#\#\# 2. Recall Relevant Medical Knowledge:
- {{Retrieve information related to the disease, medication, or procedure.}}

- {{Consider anatomy, physiology, pathology, pharmacology, and current medical guidelines.}}

\#\#\# 3. Analyze Medical Information:
- {{Combine 1. understanding the question and 2. relevant medical knowledge to connect the issue with pertinent medical knowledge using clinical reasoning.}}

- {{Consider possible explanations, mechanisms, or interactions.}}

\#\#\# 4. Assess Impacts and Considerations:

- {{Evaluate any risks, side effects, or contraindications.}}

- {{Consider specific patient factors (age, comorbidities, allergies).}}

\#\#\# 5. Provide Additional Relevant Information:

- {{Include important details that help in understanding.}}

- {{Mention any exceptions, alternative options, or preventive measures.}}

\#\#\# 6. Suggest Follow-Up Steps or Actions:

- {{If necessary, recommend consulting a healthcare professional.}}

- {{Advise on monitoring, follow-up, or further evaluation.}}

\#\#\# 7. Reference Reliable Sources:

- {{Base responses on evidence from authoritative medical texts or guidelines.}}

- {{Cite clinical studies, professional organizations, or regulatory agency information.}}

\#\#\# 8.Long-Form Answer:

- {{Combine the above reasoning to accurately and comprehensively answer the question. Provide a "long-form answer" that contains 400-500 words. The word count must not be less than 400 words.}} 

\#\#\# END

-{{Please end the output here.}}

\end{userquery}
\caption{Medical Structured Output Prompt (part2).}
\label{fig: system_prompt_2}
\end{figure*}

\begin{figure*}[ht]
    \fontsize{10.5}{12}\selectfont

    \begin{userquery}
    \textbf{Medical Structured Output Prompt (one-shot)}: 
    
Please refer to the following questions, along with examples of chain of thought and long-form answers.

Question: What is the relationship between Noonan syndrome and polycystic renal disease?

Chain of Thought:

1. Understand the Question: Noonan syndrome is a genetic disorder characterized by distinct facial features, short stature, heart defects, and developmental delays. It affects multiple systems in the body, including the cardiovascular, musculoskeletal, and endocrine systems. Polycystic renal disease, particularly autosomal dominant polycystic kidney disease (ADPKD), is a genetic condition leading to the formation of numerous cysts in the kidneys, resulting in kidney enlargement and impaired function. The question seeks to explore the potential relationship between these two conditions, particularly any shared genetic or pathological mechanisms.

2. Recall Relevant Medical Knowledge: Noonan syndrome is primarily caused by mutations in genes involved in the RAS-MAPK signaling pathway, particularly the PTPN11 gene. It affects approximately 1 in 1,000 to 1 in 2,500 births. On the other hand, polycystic renal disease is commonly caused by mutations in the PKD1 or PKD2 genes. ADPKD has a prevalence of about 1 in 400 to 1 in 1,000. Understanding the genetic basis and clinical manifestations of both conditions is crucial for identifying potential links between them.

3. Analyze Medical Information: The relationship between Noonan syndrome and polycystic renal disease may stem from shared genetic pathways or phenotypic associations. Some studies suggest that patients with Noonan syndrome exhibit renal anomalies, including renal agenesis or structural abnormalities, although true polycystic kidney disease is less commonly reported. This indicates a potential overlap in genetic vulnerabilities that could lead to renal pathologies in Noonan syndrome patients. The mechanisms may involve disruptions in cellular signaling pathways that are pivotal for kidney development and function.

4. Assess Impacts and Considerations: Patients with Noonan syndrome may have additional comorbidities that can influence renal health, such as hypertension or congenital heart defects, which could complicate the presentation of renal disease. Conversely, individuals with polycystic kidney disease are at risk of hypertension and kidney failure, potentially impacting their overall health and necessitating careful monitoring. Genetic counseling may be beneficial for families with a history of either condition to better understand the risks and implications of genetic inheritance.

5. Provide Additional Relevant Information: While there is limited direct evidence linking Noonan syndrome and ADPKD, awareness of renal complications in Noonan syndrome patients is important for clinicians. Furthermore, certain genetic syndromes may predispose individuals to multiple anomalies, making regular screenings for renal function essential in affected individuals.

6. Suggest Follow-Up Steps or Actions: For individuals diagnosed with Noonan syndrome, it is advisable to perform regular renal assessments, including ultrasound examinations to check for any renal structural anomalies. Genetic counseling can provide insights into the risks of polycystic kidney disease and the implications for family planning. Patients should be educated on signs of kidney dysfunction, such as changes in urination patterns, hypertension, or abdominal pain.

7. Reference Reliable Sources: Sources for this information include clinical guidelines from the National Kidney Foundation, the American Academy of Pediatrics, and recent genetic studies published in peer-reviewed journals regarding the genetics of Noonan syndrome and polycystic kidney disease.

8. Long-Form Answer: Noonan's syndrome is an eponymic designation that has been used during the last 8 years to describe a variable constellation of somatic and visceral congenital anomalies, which includes groups of patients previously referred to as male Turner's, female pseudo-Turner's and Bonnevie-Ullrich syndromes. It is now recognized that both sexes may show the stigmas of this condition and, unlike Turner's syndrome, there is no karyotype abnormality although there is often a familial pattern. The most commonly observed anomalies include webbing of the neck, hypertelorism, a shield-shaped chest and short stature. Congenital heart disease, principally pulmonary stenosis, and sexual infantilism often with cryptorchidism in the male subject are additional associated anomalies in this syndrome. Renal anomalies have been described rarely and usually consist of rotational errors, duplications and hydronephrosis. We report the first case of an infant who displayed many of the stigmas of Noonan's syndrome and also showed early evidence of frank renal failure secondary to renal dysplasia with cystic disease.

END
\end{userquery}
\caption{Text Generation Prompt of Med-SoCoT (one-shot).}
\label{fig: system_prompt_3}
\end{figure*}

\end{document}